\documentclass{article}
\usepackage{spconf,amsmath,graphicx}

\title{AUTOMATIC LIVER LESION SEGMENTATION USING A DEEP CONVOLUTIONAL NEURAL NETWORK METHOD}
\name{Xiao Han}
\address{Elekta Inc. \\ St. Louis, MO, USA}
\begin{document}
%
\maketitle
\begin{abstract}
Liver lesion segmentation is an important step for liver cancer diagnosis, treatment planning and treatment evaluation. LiTS (Liver Tumor Segmentation Challenge) provides a common testbed for comparing different automatic liver lesion segmentation methods. We participate in this challenge by developing a deep convolutional neural network (DCNN) method. The particular DCNN model works in 2.5D in that it takes a stack of adjacent slices as input and produces the segmentation map corresponding to the center slice. The model has 32 layers in total and makes use of both long range concatenation connections of U-Net~\cite{UNet} and short-range residual connections from ResNet~\cite{ResNet}. The model was trained using the 130 LiTS training datasets and achieved an average Dice score of 0.67 when evaluated on the 70 test CT scans, which ranked first for the LiTS challenge at the time of the ISBI 2017 conference. 
\end{abstract}
\section{Introduction}
\label{sec:intro}

Liver cancer is among the top three most deadly cancers in the modern world and contrast-enhanced computed tomography (CT) is the most commonly used modality for liver cancer screening. Segmentation of liver lesions can help oncologists diagnose the cancer, determine treatment options, and evaluate the effectiveness of cancer treatments. Manual segmentation of liver lesions from a 3D CT image is very time-consuming and prone to inter- and intra-rater variations. Therefore, automatic segmentation methods are highly desirable in clinical practice.   

Automatic liver lesion segmentation is a very challenging problem due to significant variations in location, size, shape, intensity, texture, and the number of occurrences of lesions across different patients. In addition, CT images usually have low soft-tissue contrast and  suffer from noise and other artifacts. Existing lesion segmentation methods based on intensity clustering, region growing, or deformable models have shown limited success in solving this difficult problem.

Recent developments of deep learning have revolutionized the field of artificial intelligence. Deep learning algorithms, especially deep convolutional neural networks (DCNN), have rapidly become a popular methodology for processing medical images as well, including lesion detection, segmentation, and classification~\cite{litjens2017}. For example, Christ et al.~\cite{Christ2016} have designed a two-step U-Net approach for automatic liver lesion segmentation, and reported a very high accuracy (Dice score above 0.94) on their test data. 

In this work, we also aim to exploit the advancements in deep learning and have designed another DCNN model for fully automatic liver lesion segmentation. There are two major considerations in the design of the final model. First of all, we would like to combine important features of two very successful DCNN architectures proposed in the field: the U-Net~\cite{UNet} and the ResNet~\cite{ResNet}. Second, considering the limitation of available GPU memory and limited training data, we design the DCNN model in 2.5D: the input to the model consists of several adjacent axial slices and the output is a 2D segmentation map corresponding to the center slice of the input stack. Even though a 3D DCNN model may be a more natural choice for segmenting 3D images, model capacity and input image size are restricted by available GPU memory.  We believe it is more beneficial to use a larger 2D context, and a few adjacent slices can provide sufficient complementary information in the third dimension. This is especially the case since CT images usually have much coarser resolution in the $z$-direction.

\section{DATASET AND PREPROCESSING}
\label{sec:dataset}

We used solely the LiTS datasets for the training and validation of the proposed DCNN model. The LiTS datasets consist of 200 contrast-enhanced 3D abdominal CT scans from several different clinical sites with different scanners and protocols.  The datasets thus have largely varying spatial resolution and fields-of-view. The in-plane resolution ranges from 0.60 mm to 0.98 mm, and the slice spacing from 0.45 mm to 5.0 mm. The axial slices of all scans have an identical size of $512 \times 512$, but the number of slices in each scan differs greatly and varies between 42 and 1026. 

No special pre-processing was performed except that we truncated the image intensity values of all scans to the range of $[-200, 200]$ HU to ignore irrelevant image details. Since many of the data have a large number of slices, we trained two DCNN models to reduce the overall computation time. The first one aimed to get a quick but coarse segmentation of the liver, and the second model focused on the liver region to get more detailed segmentation of the live and live lesions. To train the first model, all CT scans were re-sampled to a fixed, coarse resolution of $1 \times 1 \times 2.5$ ${\rm mm}^3$. The second model was trained using the original image resolution to avoid possible blurring from image re-sampling and to avoid missing small lesions. The training data for the second model were collected using only slices belonging to the liver region so as to focus the training on the liver and liver lesions.

\section{Method}
\label{sec:method}

The DCNN model we designed (cf. Fig.~\ref{fig:model}) belongs to the category of ``fully convolutional neural networks'' (FCNs)~\cite{long2015}. FCNs allow complete segmentation of an entire (2D) image in a single pass instead of classifying the center pixel of a small image patch each time. In addition to being more efficient, using an entire image as input offers much richer contextual information than small image patches, which usually leads to more reliable and more accurate segmentation results. 

Different FCN architectures have been proposed in the literature~\cite{long2015,noh2015,UNet,Segnet}. They typically consist of an encoding part and a decoding part. The encoding part resembles a traditional CNN that extracts a hierarchy of image features from low to high complexity. The decoding part then transforms the features and reconstructs the segmentation label map from coarse to fine resolution. A notable invention is the U-Net architecture~\cite{UNet} that introduced long range connections across the encoding part and the decoding part so that high resolution features from the encoding part can be used as extra inputs for the convolutional layers in the decoding part. This design makes it easier for the decoding part to generate high resolution predictions. These short-cuts also make the model more flexible. For example, the model can automatically learn to skip coarse level features (at bottom of the network) if high resolution features (at top of the network) are sufficient to produce accurate segmentation results.  Later, Milletari et al.~\cite{milletari2016} extended the U-Net architecture to 3D and proposed to use ResNet-like~\cite{ResNet} residual blocks as the building blocks.

\subsection{DCNN Model}
\label{ssec:model}

\begin{figure}[htbp]
  \centering
  \centerline{\includegraphics[width=8.5cm]{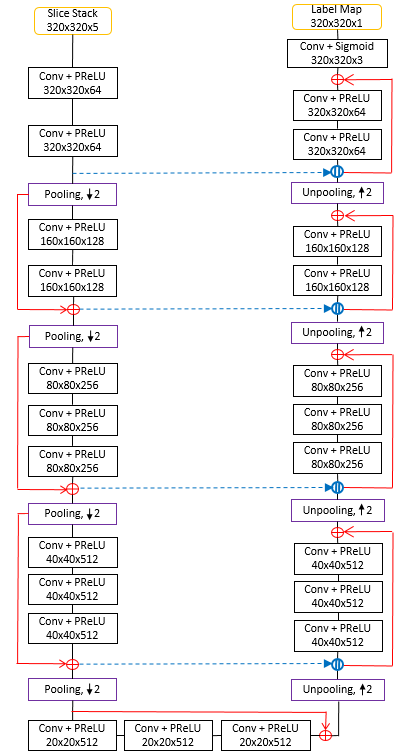}}
%
\caption{DCNN model architecture for liver and liver lesion segmentation. All convolutional layers use a kernel size of $3 \times 3$. The numbers, e.g., $160 \times 160 \times 128$,  indicate the spatial dimension and the number of channels of each convolutional layer. The red lines indicate the short-range residue connections and the blue lines indicate the long range U-Net copying and concatenation connections.}
\label{fig:model}
\end{figure}

The model we propose here is similar to that of Milletari et al.~\cite{milletari2016} in principle, where we also use both long-range U-Net and short-range ResNet skip connections (residue connections), as shown in Fig.~\ref{fig:model}. The residual connections help promote information propagation both forward and backward through the network, and improve model convergence and performance. But we design the model to work in 2.5D instead of 3D, as mentioned earlier. 

The model input is a stack of adjacent axial slices (5 slices in this work), providing large image content in the axial plane and extra contextual information in the orthogonal direction. The model output is a segmentation map corresponding to the center slice of the stack. In addition to larger input size, more layers and a much larger number of feature channels can be used in each layer than a 3D model. All convolutional layers use a filter size of $3 \times 3$ and use the {\it P}arametric {\it Re}ctified {\it L}inear {\it U}nit (PReLU) as the nonlinear activation function \cite{he2015}. The spatial size and the number of channels of the output feature maps of each convolutional layer are shown in Fig.~\ref{fig:model}.

The DCNN model represents an end-to-end mapping from the input image (slices) to a segmentation map. The model parameters are learnt from training data by minimizing a loss function. We use a weighted cross-entropy loss as the loss function in this work:
\begin{equation}
L = -\frac{1}{N} \sum_{i=1}^N \sum_{c=1}^3 w^c_i y_i^c \log{P^c_i},
\end{equation}
\noindent where $P_i^c$ denotes the predicted probability of voxel $i$ belonging to class $c$ (background, liver, or liver lesion), $y^c_i$ denotes the ground truth probability, and $w^c_i$ denotes a class-dependent weighting factor. Empirically, we set the weights to be 0.2 for background, 1.2 for liver, and 2.2 for liver lesion.  
 
As mentioned earlier, two models (with the same architecture as Fig.~\ref{fig:model}) were trained using the LiTS data. The first (liver segmentation) model was trained using all slices of the training data, but each CT image and the corresponding ground truth segmentation were first re-sampled to $1 \times 1 \times 2.5$ ${\rm mm}^3$ resolution. In addition, the lesion and liver labels were merged as a single liver label. The second model was trained using only slices inside the liver region, using the original image resolution of each scan. The second model used all three labels: background, liver, and liver lesion.

Given a new test image, the image is re-sampled to  $1 \times 1 \times 2.5$ ${\rm mm}^3$  image resolution and the first model is applied to segment each slice (using the slice and its adjacent slices as input) and generate a liver segmentation map. Once all slices of the scan are segmented, a 3D connected component labeling step is performed, and the largest connected component of all liver-labeled voxels is kept to define the initial liver segmentation. The second model is then applied to re-process each slice within the detected liver region using the original image resolution to get more detailed segmentation of the liver and the liver lesions. The refined liver and liver lesions are first merged, and a connected component labeling is run again to find the final liver region. The liver lesion segmentation result is then generated by grouping all voxels labeled as liver lesion inside the refined liver region.

The proposed DCNN model shares a major benefit as other FCN models in that the input image size at model deployment does not need to be the same as the input size during model training, because all layers act as  convolutional filters. In particular, the model can handle much larger input size during deployment since less GPU memory is needed at model testing than training.  As shown in Fig.~\ref{fig:model}, during training, we use an input size of $320 \times 320 $ ($\times 5$), but the model is applied directly to a stack of slices of size $480 \times 480$ each time and generate an output of size $480 \times 480$.
 
\subsection{Implementation Details}
\label{ssec:implem}

The DCNN model is implemented using the publicly available {\it Caffe} package \cite{Caffe}. Both models were trained from scratch using the stochastic gradient descent with momentum optimization algorithm implemented in Caffe. The initial learning rate was set to 0.001 and multiplied by 0.9 after each epoch. Each model was trained for 50 epochs.  The weight decay was set to be 0.0005 and the momentum parameter was set to 0.9.  

Simple data augmentation was performed during model training, where a random $320 \times 320 \times 5$ subregion was cropped from each training sample, and a left-right flipping was also applied randomly. To accelerate training, batch normalization is performed after each convolutional layer to reduce internal covariate shift~\cite{Ioffe2015}. 

Training of each model took about 4 days using a single NVIDIA Titan X GPU with 3584 cores and 12 GB memory. Applying the model took about 0.2 second to generate the segmentation result for each slice.  The total processing time for final lesion segmentation thus depends on the image resolution and the number of slices for each scan, which ranged from 30 seconds to 100 seconds for the LiTS test data.

\section{POSTPROCESSING}
\label{sec:postproc}

In addition to the connected component labeling step mentioned in Section~\ref{ssec:model}, we add an extra small post-processing step to help reduce false positives in the final lesion segmentation. In this step, we perform another 3D connected component labeling of all voxels labeled as lesion. We then check the maximal lesion probability (given by the DCNN model output) of all voxels in each component. If the maximal probability value is less than a threshold, the whole component is removed from the final lesion segmentation. The threshold is currently set empirically at 0.80 to generate the final results we submitted for the LiTS competition.  Based on the results provided by the organizers, our method achieved an overall average Dice value of 0.67, as shown in Table~\ref{tab:result}.

\begin{table}[]
\centering
\caption{Average segmentation accuracy on LiTS validation data.}
\label{tab:result}
\begin{tabular}{|c|c|c|c|c|}
\hline
 {\rm Dice} & \shortstack {\rm Volume \\ Overlap \\ Error}  & \shortstack {\rm Relative \\ Volume \\ Difference}  & \shortstack {\rm Average \\ Symmetric \\ Surface \\ Distance } & \shortstack {\rm Maximum \\ Symmetric \\ Surface \\ Distance } \\
\hline
0.670 & 0.450  & 0.040 & 6.660  & 57.930 \\
\hline
\end{tabular}
\end{table}

\section{DISCUSSION}
\label{sec:discussion}

The major benefit of the DCNN framework is that the training is completely 
end-to-end, using pretty much the original image data. There is no need to manually design image features and no pre-processing is necessary. Due to the large model capacity, the accuracy of DCNN model is also expected to grow with more training data. 

On the only hand, DCNNs have a very flexible architecture and different network design clearly matters. Unfortunately, there is no clear guide about what the optimal network architecture would be for a given application. Due to the long training time, it is very time consuming to evaluate different options. We thus participate in the LiTS challenge, which allows researchers to compare different designs using a common set of data. 

The lesion segmentation accuracy is still rather low, and further improvements are clearly needed. In addition to better network architectures, we plan to investigate other post-processing strategies including the 3D conditional random field method used in~\cite{CRF} and possibly using a cascade or an ensemble of DCNNs.

\bibliographystyle{IEEEbib}
\bibliography{xhan}

\end{document}